\documentclass[graybox]{svmult}

\usepackage{mathptmx}       
\usepackage{helvet}         
\usepackage{courier}        
\usepackage{type1cm}        
\usepackage{subfigure}       

\usepackage{makeidx}         
\usepackage{graphicx}        
                        
\usepackage{multicol}        
\usepackage[bottom]{footmisc}
\usepackage{url}

\makeindex             

\begin{document}

\title*{Multiple Moving Object Recognitions in video based on Log Gabor-PCA Approach}
\author{M. T Gopalakrishna, M. Ravishankar and D. R Rameshbabu}
\institute{M. T Gopalakrishna \at Department of ISE, Dayananda Sagar College of Engineering, Shavige Malleshwara Hills, Bangalore \email{gopalmtm@gmail.com},
\and M. Ravishankar \at Department of ISE, Dayananda Sagar College of Engineering, Shavige Malleshwara Hills, Bangalore \email{ravishankarmcn@gmail.com}
\and D. R Rameshbabu \at Department of CSE, Dayananda Sagar College of Engineering, Shavige Malleshwara Hills, Bangalore \email{bobrammysore@gmail.com}}
%

\maketitle
\abstract*{Object recognition in the video sequence or images is one of the sub-field of computer vision. Moving object recognition from a video sequence is an appealing topic with applications in various areas such as airport safety, intrusion surveillance, video monitoring, intelligent highway, etc. Moving object recognition is the most challenging task in intelligent video surveillance system. In this regard, many techniques have been proposed based on different methods. Despite of its importance, moving object recognition in complex environments is still far from being completely solved for low resolution videos, foggy videos, and also dim video sequences. All in all, these make it necessary to develop exceedingly robust techniques. This paper introduces multiple moving object recognition in the video sequence based on LoG Gabor-PCA approach and Angle based distance Similarity measures techniques used to recognize the object as a human, vehicle etc. Number of experiments are conducted for indoor and outdoor video sequences of standard datasets and also our own collection of video sequences comprising of partial night vision video sequences. Experimental results show that our proposed approach achieves an excellent recognition rate. Results obtained are satisfactory and competent.}

\abstract{Object recognition in the video sequence or images is one of the sub-field of computer vision. Moving object recognition from a video sequence is an appealing topic with applications in various areas such as airport safety, intrusion surveillance, video monitoring, intelligent highway, etc. Moving object recognition is the most challenging task in intelligent video surveillance system. In this regard, many techniques have been proposed based on different methods. Despite of its importance, moving object recognition in complex environments is still far from being completely solved for low resolution videos, foggy videos, and also dim video sequences. All in all, these make it necessary to develop exceedingly robust techniques. This paper introduces multiple moving object recognition in the video sequence based on LoG Gabor-PCA approach and Angle based distance Similarity measures techniques used to recognize the object as a human, vehicle etc. Number of experiments are conducted for indoor and outdoor video sequences of standard datasets and also our own collection of video sequences comprising of partial night vision video sequences. Experimental results show that our proposed approach achieves an excellent recognition rate. Results obtained are satisfactory and competent.}


\section{Introductions}
\label{sec:1}
Moving object recognition from a video sequence is a fascinating topic with applications in various areas such as airport safety, intrusion surveillance, video monitoring, intelligent highway, etc \cite{1}.  Moving object recognition in the video sequence has been considered as one of the most fascinating and challenging areas in computer vision and pattern recognition, in recent years. In last one decade, researchers have proposed a variety of approaches for moving object detection and classification, most of them are based on motion and shape features. For example, an end-to-end method for extracting moving targets from a real-time video stream has been presented by \cite{2}.  This method is applicable to human and vehicle classification with shapes that are remarkably different. Petrovic et al \cite{3} extract gradient features from reference patches in images of car fronts, and recognition is performed in two stages. Gradient-based feature vectors are used to produce a ranked list of possible classes of the candidate. A novel match refinement algorithm is used to refine the obtained result. There are many other moving objects classification methods based on multi-feature fusion \cite{4, 5, 6, 7, 8}. Image features are arguably the most fundamental task in moving object recognition. Generally, there are two categories of feature representation: appearance feature based and geometric feature based. Appearance features have been demonstrated to be better than geometric features, for the superior insensitivity to noises, especially illumination changes, foggy weathers etc. Gabor wavelets are reasonable models of visual processing in primary visual cortex and are one of the most successful approaches to describe local appearance of the human face, exhibiting powerful characteristics of spatial locality, scale and orientation selectivity \cite{9}. However, they fail to provide excellent simultaneous localization of the spatial and frequency information due to the constraints of the narrow spectral bandwidth, which is crucial to the analysis of highly complex scene. Jamie Cook et al. \cite{10} proposed a system for 3D Face Recognition using Log-Gabor Filter to obtain a simultaneous response that is Gaussian when viewed on a logarithmic frequency scale instead of a linear one. As an alternative to traditional common approaches, D. Field \cite{11} proposed a Log Gabor Filter to perform DC compensation and to overcome the bandwidth limitations of traditional Gabor Filter banks. However, the dimensionality of the resulting data is exceptionally high. For this reason, a computationally effective approach is needed. One common choice would be Principle Component Analysis (PCA). As per the review reports, it is clear that research on moving object recognitions in complex background, low contrast, foggy videos and cluttered are still a challenging problem. Motivated by the above facts, in the present work, an idea of LoG-Gabor-PCA approach is explored to obtain desirable high pass characteristics as well to capture more information in high frequency areas for efficient moving objects recognitions in the video sequences.

\section{Proposed Method}
\label{sec:2}
\begin{figure}[h!]
\centering
\includegraphics[height = 2 in, width = 3.7 in]{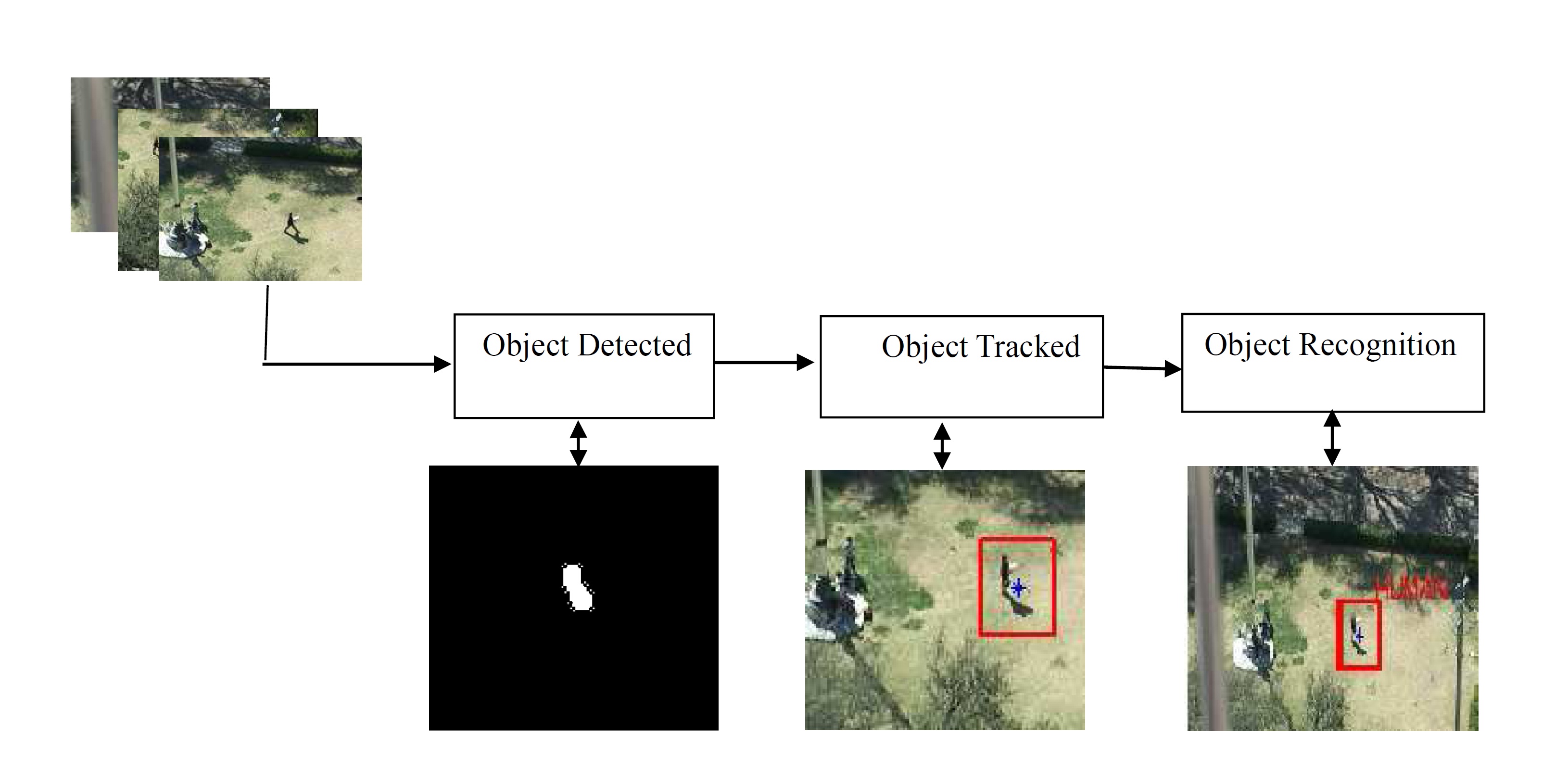}

\caption{Block Diagram of the Proposed System.}
\label{blk_1}      
\end{figure}

The proposed system comprises of three main steps. The primary step is moving object detection by using Tensor Locality Preserving Projections (Ten-LoPP). The second step is moving object tracking based on the Centroid and Area of detected object. Finally, moving object recognition is performed by using Log-Gabor-PCA approach. The entire processing of the proposed system is illustrated in Fig.~\ref{blk_1}.

\subsection{Moving object Detection and Tracking}
\label{subsec:2}
\begin{figure*}
\centering
\subfigure[Successfull detection of moving object]{
\includegraphics[width=1.7 in, height=1 in]{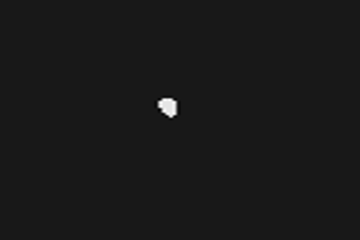}
\label{det_2}
}
\subfigure[Moving Object Tracking]{
\includegraphics[width=1.7 in, height=1.1in]{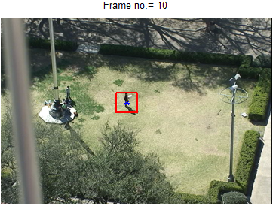}
\label{track_3}
}
\caption[]{Results of Detection and Tracking}
\label{dt_tr2}
\end{figure*}

In the proposed method of moving object detection and tracking, the method proposed by us in system \cite{12} is considered. Here, moving object detection is performed by using Ten-LoPP and moving object tracking is done by considering the Centroid and Area of detected object. Results of Detection and Tracking is shown in Fig.~\ref{det_2}, \& ~\ref{track_3}. In this work concentration is mainly given towards recognitions of multiple moving object in video sequences and hence recognitions steps described in details.

\subsection{Moving Object Recognition}
\label{subsec:3}
This section explains the process of moving object recognition in video sequences. The detected moving object features are extracted by applying LoG-Gabor filter. The feature dimension reduction is performed using PCA and are stored in the library.

\subsubsection{LoG-Gabor}
\label{subsubsec:4}
In this section, the reason of applying Log-Gabor filter for the moving object recognition is illustrated. Excellent spatial and frequency information is provided by the Gabor filters for the object localization in scenes, and the description is given about the characteristics of scale, orientation and spatial locality selectivity \cite{11}. The main drawback of Gabor filter is that it doesn't provide the excellent simultaneous localization of the spatial and frequency information and its limitation being that the maximum bandwidth captured by a Gabor filter cannot exceed approximately to one octave. This drawback can be overcome by using Log-Gabor function proposed by field \cite{11}. It is possible to vary the bandwidth form one to three octaves using Log-Gabor filters. Log-Gabor filters have features like null DC component, which reinforces the contrast ridges and edges of images and can be constructed with an arbitrary bandwidth which can be optimized to produce filter with minimal spatial extent. In linear frequency scale, co-ordinates by the transfer function H (f, $\theta $) in the Log-Gabor filters defined in the frequency domain using polar can be represented in a polar form as

\begin{equation} 
H\left(f,\theta \right)=H_fXH_{\theta }=exp\left\{\frac{1}{2\ }\frac{{\left(ln\frac{f}{f_0}\right)}^2}{{\left(ln\frac{{\sigma }_f}{f}\right)}^2}\right\}\ exp\left\{-\frac{1}{2\ }\frac{{\left(\theta -{\theta }_0\right)}^2}{{{\sigma }_{\theta }}^2}\right\}\  
\label{5lgf_1}
\end{equation} 

\begin{figure*}
\centering
\subfigure[]{
\includegraphics[width=1.5 in, height=1in]{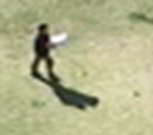}
\label{5lgff_2}
}
\subfigure[]{
\includegraphics[width=1.5 in, height=1in ]{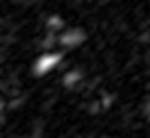}
\label{5lgff_3}
}
\caption[]{Results of Gabor filtering process:(a) original image.(b) Log Gabor filtered Image.}
\label{5lgf_33}
\end{figure*}

The radial component H${}_{f}$ controlling the bandwidth and the angular component H${}_{\theta }$ controlling the spatial orientation that the filter responds \cite{13}. The resultant image of applied LoG Gabor filter is shown in Fig.~\ref{5lgff_3}. The obtained LoG Gabor filter features whose dimensionality is exceptionally high. For this reason, one of the computationally effective subspace approach is needed. Which is Principle Component Analysis (PCA) have been used which reduce the dimensions space significantly.

\subsubsection{Combination of LoG Gabor-PCA}
\label{subsubsec:5}
\begin{figure*}
\centering
\subfigure[]{
\includegraphics[width=1.7 in, height=1.1 in]{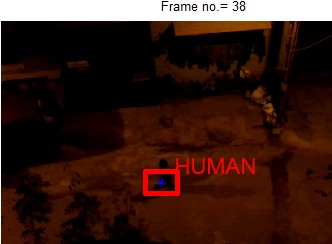}
\label{5lgfr_1}
}
\subfigure[]{
\includegraphics[width=1.7 in, height=1in]{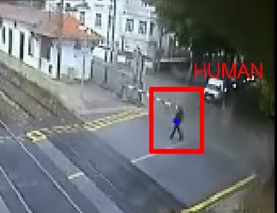}
\label{5lgfr_2}
}
\caption{Results of Moving Objects Recognition in test video sequences.}
\label{5lgf_rf5}
\end{figure*}

The accuracy is significantly high for the object recognition from Log-Gabor filter based PCA for the standard and our own collected Datasets. This is capable of recognizing objects in video with small far captured objects, diffuse glow effect, dark objects, dark backgrounds etc. The construction of Log-Gabor filters are done with arbitrary bandwidth and the optimization can be done to produce filters with minimal spatial extent. There are two prominent features: First, Log-Gabor functions always have no DC component, and second, the transfer function of the Log-Gabor function has an extended tail at high frequency end \cite{11}.\\
Feature vectors of moving objects are extracted for a given set of training images through Log-Gabor approach. For finding the Principle Components and to reduce the dimensional space of the image to store in the library the reduced image data is further processed by PCA. Feature vectors are extracted through appropriate Log-Gabor approach for testing purpose of moving object image sequence. Further reduction of the dimensional space of the image is done and features are extracted using PCA, and the better classification of the moving object is done by measuring the angle based distance of mean values of training image sequences in each class and the testing image sequences. The recognition results obtained are shown in Fig.~\ref{5lgf_rf5}.

\section{Experimental Results and Comparative Study}
\begin{figure*}
\centering
\subfigure[]{
\includegraphics[height = 1 in, width = 1 in]{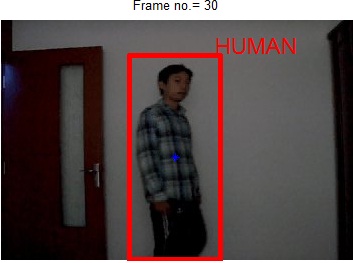}
\label{5sdr1}
}
\subfigure[]{
\includegraphics[height = 1 in, width = 1 in]{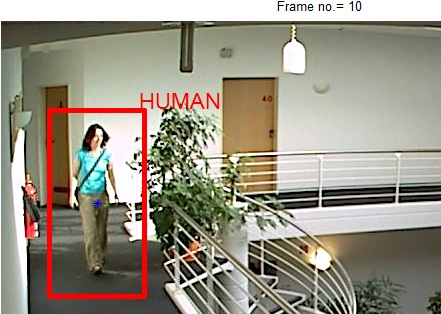}
\label{5sdr2}
}
\subfigure[]{
\includegraphics[height =  1 in, width = 1 in]{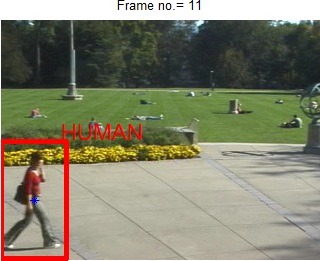}
\label{5sdr3}
}
\subfigure[]{
\includegraphics[height =  1 in, width = 1 in]{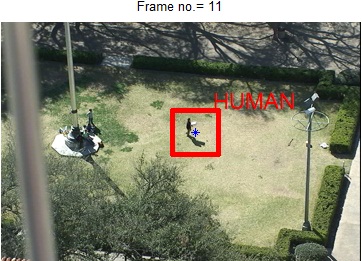}
\label{5sdr4}
}
\subfigure[]{
\includegraphics[height =  1in, width = 1 in]{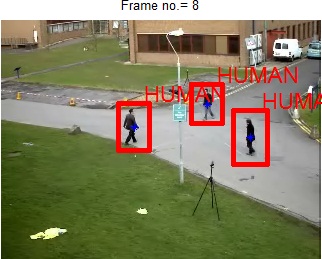}
\label{5sdr5}
}
\subfigure[]{
\includegraphics[height =  1 in, width = 1 in]{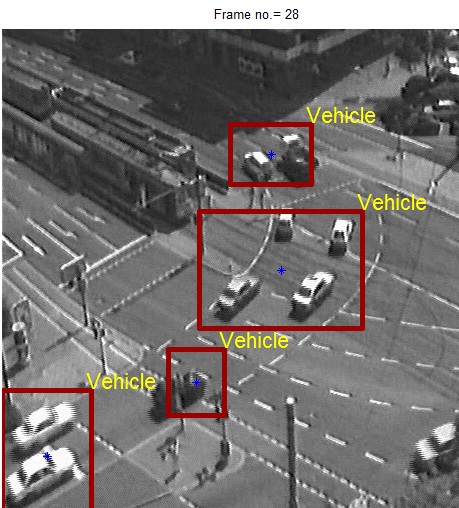}
\label{5sdr6}
}
\subfigure[]{
\includegraphics[height =  1 in, width = 1 in]{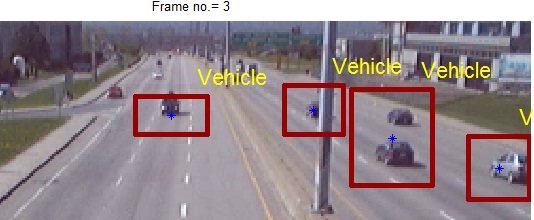}
\label{5sdr7}
}
\subfigure[]{
\includegraphics[height = 1 in, width = 1 in]{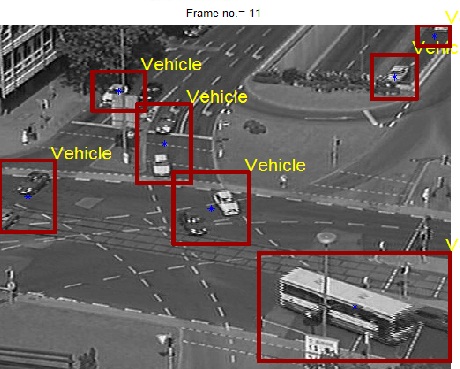}
\label{5sdr8}
}
\subfigure[]{
\includegraphics[height =  1 in, width = 1 in]{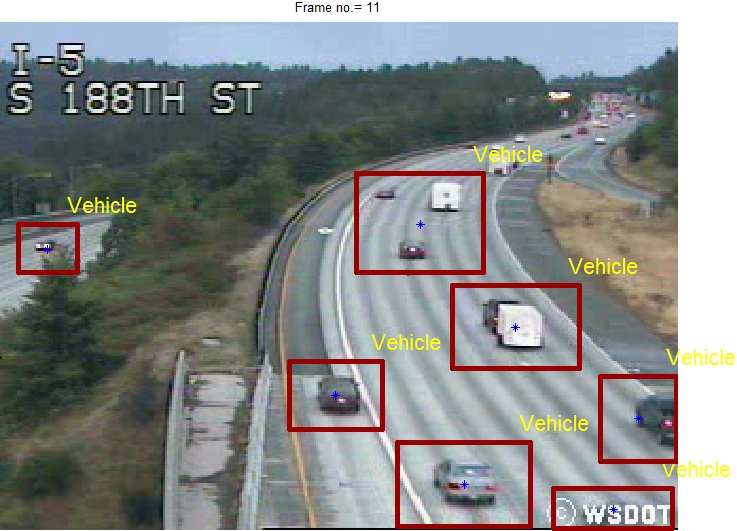}
\label{5sdr9}
}
\subfigure[]{
\includegraphics[height =  1 in, width = 1 in]{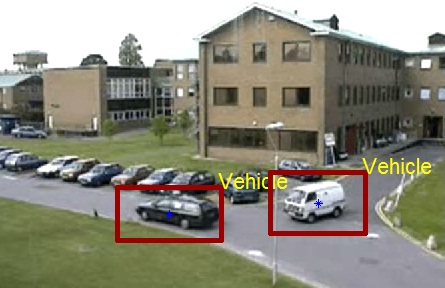}
\label{5sdr10}
}
\subfigure[]{
\includegraphics[height =  1 in, width = 1in]{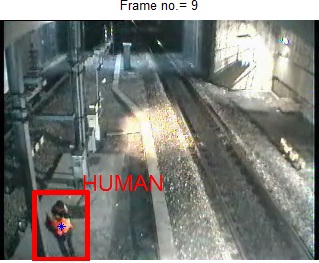}
\label{5sdr11}
}\subfigure[]{
\includegraphics[height =  1 in, width = 1in]{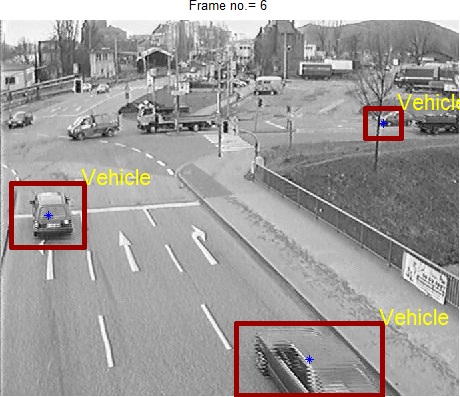}
\label{5sdr12}
}
\caption{Results of Moving Object Recognition of standard Video Sequences using proposed LoG-Gabor-PCA approach}
\label{5sdar1}
\end{figure*}

\begin{figure*}
\centering
\subfigure[]{
\includegraphics[height =  1 in, width = 1 in]{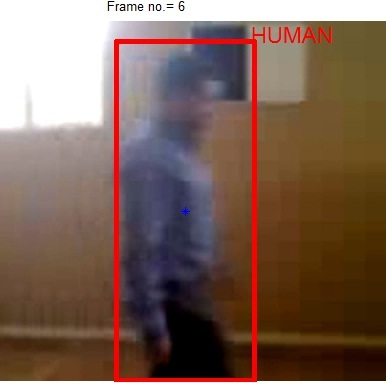}
\label{5owrr1}
}\subfigure[]{
\includegraphics[height =  1 in, width =1 in]{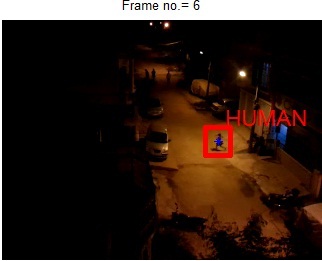}
\label{5owrr2}
}
\subfigure[]{
\includegraphics[height =  1 in, width = 1 in]{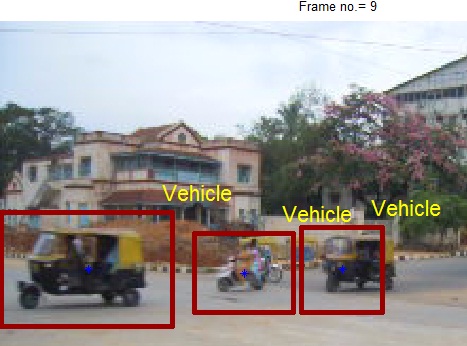}
\label{5owrr3}
}
\subfigure[]{
\includegraphics[height =  1 in, width =1 in]{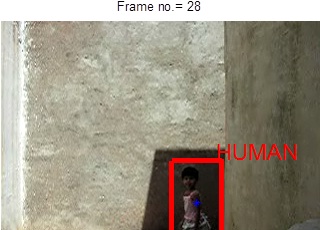}
\label{5owrr4}
}
\caption{Results of Moving Object Recognition of our own video sequences using proposed LoG-Gabor-PCA approach.}
\label{5ourown2}
\end{figure*}

\begin{figure*}
\centering
\subfigure[]{
\includegraphics[height =  1 in, width = 1.2 in]{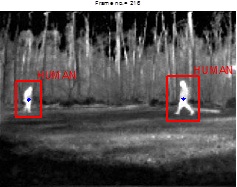}
\label{5irr1}
}
\subfigure[]{
\includegraphics[height =  1 in, width = 1.2 in]{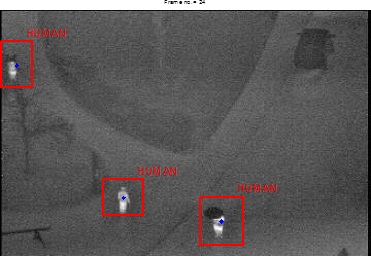}
\label{5irr2}
}\subfigure[]{
\includegraphics[height =  1 in, width = 1.2 in]{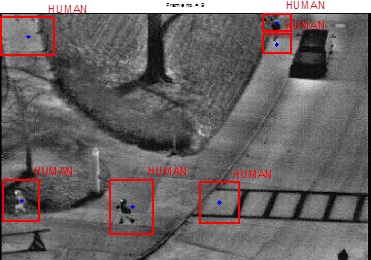}
\label{irr3}
}
\caption{Results of Moving Object Recognition of standard OTCBVS datasets (Infrared Images) Video Sequences using proposed LoG- Gabor-PCA approach.}
\label{5irr51}
\end{figure*}

The proposed system is implemented in Pentium IV 1 GHz processor with MATLAB 10. The system is experimented on standard PETS, OTCBVS, and Videoweb Activities datasets and also on our own collected video sequences comprising various environmental scenes. The proposed system capable of recognizing the moving objects of indoor and outdoor environments of standard and also of our own collected video sequences efficiently by using LoG-Gabor-PCA Approach. The Fig. ~\ref{5sdar1}, \ref{5ourown2} \& \ref{5irr51} shows successive moving object recognitions of a scene of standard dataset and our own datasets respectively. The LoG-Gabor-PCA approach is successfully experimented on standard and our own collected datasets. Recognition accuracy is also tabulated for different standard and our own collected datasets which is shown in Table~\ref{tab:1}. From the table, it is clear that the percentage of recognition accuracy 90.5

\begin{table}
\caption{Percentage of Recognition Accuracy}
\label{tab:1}       
\begin{tabular}{p{2.3cm}p{2.8cm}p{2.8cm}p{2.8cm}}
\hline\noalign{\smallskip}
Input Sequences & Correctly Recognized & Incorrectly Recognized & Recognition Accuracy  \\
\noalign{\smallskip}\svhline\noalign{\smallskip}
Arial Fig 3      & 25  & 03 &  88  \\
OTCBVS Fig 7(a)  & 43  & 07  & 86 \\
Fig 8(a)         & 23  & 02  & 92 \\
Fig 8(b)         & 24  & 01  & 96 \\
\noalign{\smallskip}\hline\noalign{\smallskip}
\end{tabular}
\end{table}
\section{conclusion}
\label{sec:3}
The Proposed Method is successfully implemented and experimented on standard and our own collected datasets of different environments. A LoG-Gabor-PCA based moving object recognition scheme has been developed to overcome the drawback of the original PCA and Gabor Filter Approaches. Applying LoG-Gabor-PCA approach on detected moving objects is a significant advantage in achieving dimensionality reduction and insensitive texture feature extraction. The LoG-Gabor-PCA approach gives better recognition accuracy in recognizing moving objects of video sequences. The performance of the moving object detection and recognition can be improved in specific circumstances such as occlusion between objects and will be considered in our future work.
%
%

\begin{thebibliography}{99.}

\bibitem{1} Hsu, and Wallace: An industrial network flow information integration model for supply chain management and intelligent transportation,  Enterprise Information Systems 1(3), pp.327-351,(2007).

\bibitem{2}	Lipton, Alan J., Hironobu Fujiyoshi, and Raju S. Patil.: Moving target classification and tracking from real-time video. In: Applications of Computer Vision , WACV'98. IEEE Proceedings, pp. 8-14,(1998).

\bibitem{3} Petrovic, V. S., and T. F. Cootes.: Vehicle type recognition with match refinement.In International Conference on Pattern Recognition,vol. 3, no. 8, pp. 95-98, (2004). 

\bibitem{4} Lin, Yingqiang, and Bir Bhanu., Evolutionary feature synthesis for object recognition.. Systems, Man, and Cybernetics, Part C: Applications and Reviews, IEEE Transactions on Vol.35, no. 2, pp. 156-171,(2005).


\bibitem{5} Sullivan, Geoffrey D., Keith D. Baker, Anthony D. Worrall, C. I. Attwood, and P. M. Remagnino.: Model-based vehicle detection and classification using orthographic approximations.Image and Vision Computing Vol.15,No. 8, pp.649-654,(1997).

\bibitem{6} Bergboer, N. H., E. O. Postma, and HJ van den Herik.: Context-based object detection in still images., Image and Vision Computing, Vol.24, No. 9, pp. 987-1000, (2006).

\bibitem{7}	Takano, Shigeru, Teruya Minamoto, and Koichi Niijima.: Moving object recognition using wavelets and learning of eigenspaces. pp.151, (1998).

\bibitem{8} Zanin, Michele, Stefano Messelodi, and Carla Maria Modena.: An efficient vehicle queue detection system based on image processing. In Image Analysis and Processing, IEEE Proceedings, pp. 232-237. (2003).

\bibitem{9} Liu, C., Wechsler, H.: Independent Component Analysis of Gabor Features for Face Recognition. IEEE Trans. Neural Networks Vol 14, No 4, pp.919-928, (2003).

\bibitem{10} Jamie Cook, Vinod Chandran, Sridha Sridharan and Clinton Fookes.: Gabor Filter Bank Representation for 3D Face Recognition, Proc. IEEE Digital Imaging Computing: Techniques and Applications, pp. 16-23, (2005).

\bibitem{11} D. Fields.: Relations between the statistics of natural images and the response properties of cortical cells, Journal of Optical Society of America, vol. 4, no. 12, pp. 2379-2394,(1987).
%
\bibitem{12} M. T. Gopala krishna, M. Ravishankar, D. R Rameshbabu.: Ten-LoPP: Tensor Locality Preserving Projections Approach for Moving Object Detection and Tracking, 9th International Conference on CIT, Springer-Advances in Intelligent Systems and Computing Volume 209, pp 291-300, (2013).

\bibitem{13} Lajevardi, S.M., Hussain, Z.M.: Facial Expression Recognition Using Log-Gabor Filters and Local Binary Pattern Operators. In Proceedings of the International Conference on Communication, Computer and Power, pp. 349-353, (2009).
\end{thebibliography}

\end{document}